\documentclass[10pt,twocolumn,letterpaper]{article}

\usepackage[pagenumbers]{cvpr} 
\usepackage{multirow}
\usepackage{graphicx} 
\usepackage{algpseudocode}
\usepackage{times}
\usepackage{epsfig}
\usepackage{graphicx}
\usepackage{amsmath}
\usepackage{amssymb}

\newcommand{\model}{FGP\xspace}
\usepackage{algorithm}

\definecolor{cvprblue}{rgb}{0.21,0.49,0.74}
\usepackage[pagebackref,breaklinks,colorlinks,allcolors=cvprblue]{hyperref}

\def\paperID{18420} 
\def\confName{CVPR}
\def\confYear{2025}

\title{FGP: Feature-Gradient-Prune for Efficient Convolutional Layer Pruning}

\author{
Qingsong Lv\textsuperscript{1}, Jiasheng Sun\textsuperscript{1}, Sheng Zhou\textsuperscript{1}$^{\dagger}$, 
Xu Zhang\textsuperscript{1}, Liangcheng Li\textsuperscript{1}, Yun Gao\textsuperscript{2}, 
\\Sun Qiao\textsuperscript{1}, Jie Song\textsuperscript{1}, Jiajun Bu\textsuperscript{1}$^{\dagger}$ \\
\textsuperscript{1}Department of Computer Science, Zhejiang University, Hangzhou, China \\
\textsuperscript{2}Kingsoft Corporation, Wuhan, China \\
{\tt\small lvqingsong.0@outlook.com, Sunjiasheng.0@outlook.com, zhousheng\_zju@zju.edu.cn,} \\ 
{\tt\small 635926955@qq.com, xml6125jhev@hotmail.com, gaoyun.0@outlook.com, sunqiao\_zju@zju.edu.cn,} \\ 
{\tt\small sjie@zju.edu.cn, bjj@zju.edu.cn}
}

\begin{document}
\maketitle
\begin{abstract}
To reduce computational overhead while maintaining model performance, model pruning techniques have been proposed. Among these, structured pruning, which removes entire convolutional channels or layers, significantly enhances computational efficiency and is compatible with hardware acceleration. However, existing pruning methods that rely solely on image features or gradients often result in the retention of redundant channels, negatively impacting inference efficiency.
To address this issue, this paper introduces a novel pruning method called Feature-Gradient Pruning (FGP). This approach integrates both feature-based and gradient-based information to more effectively evaluate the importance of channels across various target classes, enabling a more accurate identification of channels that are critical to model performance.
Experimental results demonstrate that the proposed method improves both model compactness and practicality while maintaining stable performance. Experiments conducted across multiple tasks and datasets show that FGP significantly reduces computational costs and minimizes accuracy loss compared to existing methods, highlighting its effectiveness in optimizing pruning outcomes. The source code is available at: \url{https://github.com/FGP-code/FGP}.
\end{abstract}

\section{Introduction}
Convolutional Neural Networks (CNNs) have demonstrated exceptional performance in computer vision tasks \cite{krizhevsky2012imagenet,girshick2014feature,gavrila1999real,long2015fully}, and their increasing depth and width have significantly enhanced performance across multiple scenarios. However, this increased model capacity also leads to higher computational and storage demands, hindering the deployment of CNNs on resource-constrained edge devices\cite{wu2023fibonet,yu2018bisenet}. 
To reduce computational costs while preserving model performance, model pruning techniques have emerged \cite{chen2015compressing,han2015learning,lopez2024filter}. Existing pruning methods are primarily divided into two classes: structured pruning and unstructured pruning. Compared to unstructured pruning \cite{frankle2018lottery,lee2018snip} which removes individual weights or connections, \textit{structured pruning} directly removes entire convolutional filters, channels, or layers \cite{huang2022dyrep,liu2022energy},
which not only effectively reduces computation but also retains a structure more suitable for hardware acceleration, making it more feasible in practical applications. Many studies have proposed effective structured pruning methods and achieved significant results in CNN models.

\begin{figure}[t]
    \centering
    \includegraphics[width=0.47\textwidth]{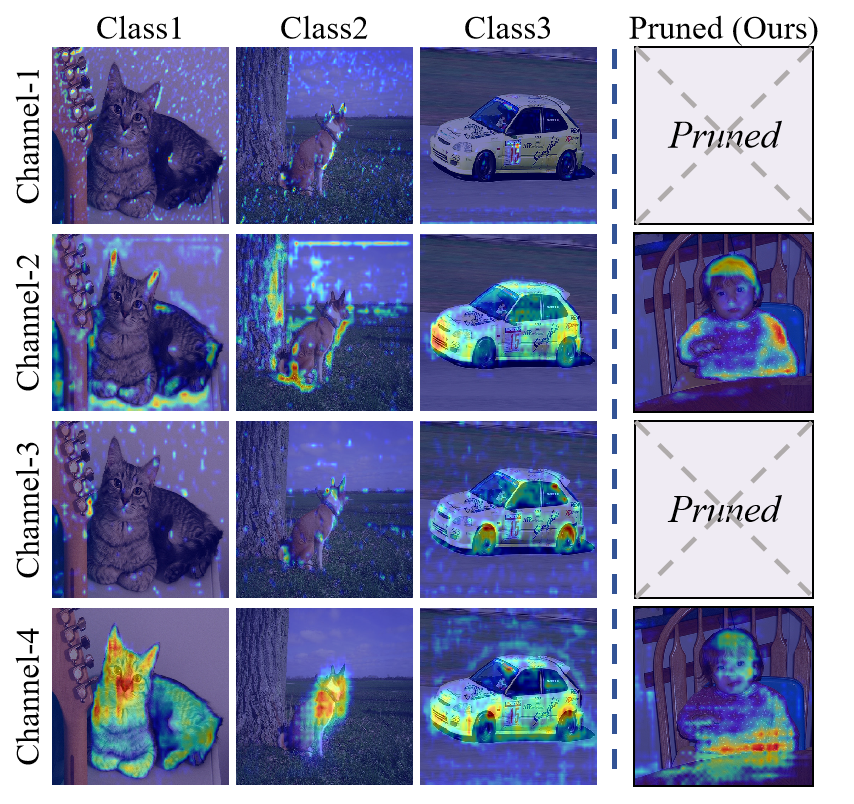}
    \caption{The visualization results show one channel from each of the four convolutional layers, along with heatmaps for three classes. \model retains Channel-2 and 4, and pruned model keeps only those channels with strong support values across all classes.} 
    \label{Fig1}
\end{figure}

While existing structured pruning methods effectively reduce parameter counts and computation, they still exhibit several limitations during pruning, affecting performance across different tasks and the model's performance after pruning.
Current structured pruning methods can be divided into two classes: feature-based pruning methods and gradient-based pruning methods, distinguished by the criteria used to evaluate channel importance.
Feature-based pruning methods typically assess the importance of channels by measuring the magnitude of convolutional layer weights (e.g., L1 norm \cite{liu2017learning}) or the $\gamma$ values in Batch Normalization layers \cite{koo2023ppt}.
These methods are computationally efficient and straightforward, yet they may fall short in capturing complex inter-channel relationships.

On the other hand, gradient-based pruning methods evaluate the contribution of channels to task optimization by analyzing gradient information (e.g., RFC \cite{he2022filter} or DMCP \cite{guo2020dmcp}). 
They reflect the actual impact of channels during the training process but often lack an overall grasp of channel characteristics, making it easy to overlook the feature representations of channels and potentially leading to an inability to maintain a well-structured feature distribution, which may affect the model's stability.
Feature-based methods provide a global perspective on overall feature data, ensuring a more balanced feature distribution across channels and reducing the risk of overlooking feature structure by focusing solely on training dynamics. In contrast, gradient-based methods select channels based on actual contributions, ensuring that retained channels are truly effective in optimizing the task and reducing redundancy. Therefore, we consider combining feature information.

Previous studies \cite{Rao2024,Vinogradova2020} have shown that the values of heatmaps generated based on gradient and feature information have a significant positive correlation with task accuracy.
Therefore, we consider using heatmaps for channel importance evaluation as an effective approach, which can help us more comprehensively assess the actual contribution of channels to different classes during the pruning process. 
Inspired by this, we propose a method called Feature-Gradient Pruning (\model). 
As shown in Fig. \ref{Fig1}, the \model method can identify channels that provide important support across all classes, while removing redundant channels that are only useful for specific classes, achieving finer-grained pruning optimization, which can mproves model compactness and practical value while maintaining high performance.

In summary, our contributions are as follows:
\begin{itemize}
    \item \model combines feature and gradient information to assess the importance of channels, addressing the shortcomings of existing pruning methods. By integrating these two types of information, \model can more accurately identify the channels that are crucial for model performance, thereby improving model efficiency.
    \item \model refines the channel selection process, making the pruning more precise. It retains channels that are important for all categories while removing redundant channels that are only useful for specific categories. This more refined pruning approach enhances the model's compactness and practicality.
    \item \model uses a Top $k$ strategy based on the concentration of support values across different classes to select channels, identifying those with the highest importance across all classes. The pruning ratio is not fixed but dynamically adjusted based on the support values of the channels. 
\end{itemize}

\section{Related Work}
\subsection{Pruning Methods}

Pruning has become a mainstream method for deep neural network compression and has received widespread attention in recent years. It improves computational and storage efficiency by removing redundant parameters while maintaining model performance. Structured pruning methods reduce the network size by removing structural units such as convolutional kernels, neurons, or channels. 

Structured pruning methods include gradient-based pruning, feature-based pruning, and a combination of gradient and feature-based pruning methods. Gradient-based pruning utilizes gradient information to determine the importance of each parameter. Niu et al. \cite{Niu2023} introduced a block pruning method that defines a sensitivity metric to calculate the gradient sensitivity relative to the input. He et al. \cite{He2023} proposed a gradual pruning technique based on gradients, incrementally increasing the pruning rate to determine the optimal subnetwork for each language pair in multilingual models. However, these methods do not consider the impact of feature selection on pruning quality.

\begin{figure*}[t]
    \centering
    \includegraphics[scale=0.49]{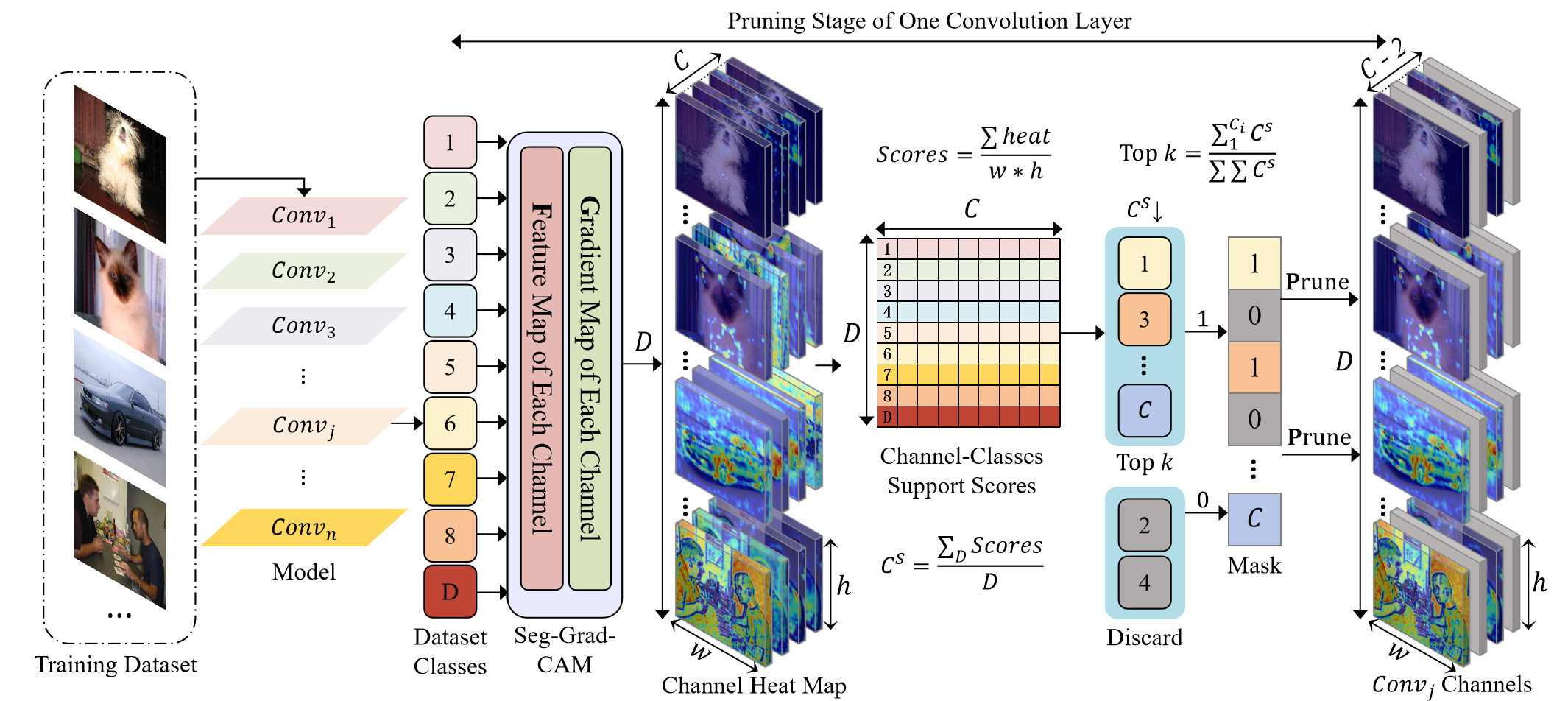}
    \caption{\model pruning framework, where $\text{Conv}_{j}$ is used as an example. The process calculates each channel’s support value across all classes in the dataset. These support values are then summed to assess the overall importance of each channel. Channels are ranked by their importance scores, and the Top $k$ channels are selected. Pruning is then applied based on these rankings, resulting in a refined, pruned set of channels for the layer. Best viewed on screen.}
    \label{Fig2}
\end{figure*}

Feature-based pruning methods screen redundant parameters based on the contribution of feature maps or channels. Dmitry Molchanov et al. \cite{Molchanov2017} proposed a variational inference-based method that gradually eliminates unimportant features to achieve pruning. Gao et al. \cite{Gao2018} introduced a dynamic pruning method that reduces feature distribution similarity through feature enhancement and suppression. Liu et al. \cite{Liu2023} proposed a pruning method based on feature similarity and 2D entropy. While these methods demonstrate good pruning effects, they do not fully account for the impact of gradients on pruning quality.

Our research aims to leverage the combined strengths of gradient and feature information to enhance pruning effectiveness. By incorporating both types of information, we seek to identify channels that play a critical role in supporting all classes, ensuring that essential features are preserved across the pruning process.

\subsection{CAM-Based Lightweighting Methods}

In neural network interpretability research, visualization techniques are essential tools for analyzing the decision-making process of models. Grad-CAM \cite{Selvaraju2017}, a classic gradient-based method, generates a weighted activation map by computing the gradient of the output class with respect to the last convolutional layer. Grad-CAM++ \cite{Chattopadhay2018} improves upon this method by addressing the limitations of Grad-CAM in specific scenarios using pixel-level gradient weighting. Seg-Grad-CAM \cite{Vinogradova2020} extends Grad-CAM to image segmentation tasks, focusing on visualizing the areas the model attends to during segmentation prediction. It more accurately explains the model's decision-making process, particularly excelling in image segmentation tasks. Furthermore, Refined Weak Slice (RWS) delves into the relationship between the heatmap values generated by Seg-Grad-CAM and task accuracy \cite{Rao2024}. By examining the correlation between segmentation accuracy and average heatmap values, which has inspired our approach to identifying important channels within the model.


\section{Preliminaries}

\subsection{Background and Notations}

Given a deep neural network model and an input image dataset \( X = \{x_1, x_2, \cdots, x_N\} \) with labels \( Y = \{y_1, y_2, \cdots, y_N\} \), where each sample belongs to one of \( K \) classes. CAM aims to provide visual explanations of the model’s decision process. CAM techniques leverage feature representations from convolutional layers to highlight image regions associated with specific classes, offering insight into the regions most relevant to the model’s predictions.

The original CAM method produces an activation map denoted as \( \text{CAM} \in \mathbb{R}^{h \times w} \), where \( h \) and \( w \) represent the height and width of the image. Various CAM methods, such as Grad-CAM, Grad-CAM++, and Score-CAM, employ different feature extraction and pooling strategies to enhance interpretability and adapt to various scenarios.

\subsection{Gradient Activation Heatmap}

\noindent \textbf{Grad-CAM.} The fundamental concept of Grad-CAM is to leverage the gradient information of the final convolutional layer to produce a class activation map. For a given class logit \( y^c \), the gradients \( \frac{\partial y^c}{\partial A^k} \) with respect to the feature map \( A^k \) are computed, producing the importance weights:

\begin{equation}
    \centering	
\alpha_k^c = \frac{1}{Z} \sum_{i} \sum_{j} \frac{\partial y^c}{\partial A_{ij}^k},
\end{equation}

\noindent where \( Z \) denotes the area of the feature map, and \( \alpha_k^c \) represents the weight of the \( k \)-th channel. The Grad-CAM heatmap is then generated by linearly combining the feature maps with the computed weights:

\begin{equation}
    \centering	
    L^c = \text{ReLU}\left( \sum_k \alpha_k^c A^k \right).
\end{equation}

The ReLU function is used to retain only the regions positively contributing to the target class, ensuring that the resulting map highlights class-specific regions.

\noindent \textbf{Grad-CAM++.} It refines the Grad-CAM weighting approach, enhancing performance in multi-object and complex visual scenes. It introduces higher-order gradient terms, allowing a more precise assessment of class-discriminative regions. The modified weight computation in Grad-CAM++ is given by:

\begin{equation}
    \centering	
\alpha_k^{c++} = \sum_{i,j} \frac{\partial^2 y^c}{\partial A_{ij}^k}.
\end{equation}

\subsection{Feature Scores Heatmap}
\noindent \textbf{Score-CAM.} It eliminates the reliance on gradient information, instead employing the activation strength of feature maps in each convolutional layer. Each feature map is weighted according to its contribution to the final class score, computed via the Softmax function:

\begin{equation}
    \centering	
\beta_k = \frac{\text{exp}(S(A^k))}{\sum_{j} \text{exp}(S(A^j))},
\end{equation}

\noindent where \( S(A^k) \) represents the classification score obtained by weighting each feature map channel. The Score-CAM activation map is expressed as follows.

\begin{equation}
    \centering	
L^{c} = \sum_k \beta_k A^k.
\end{equation}

By using score-based weighting rather than gradient dependence, Score-CAM provides a stable class activation map generation approach that maintains interpretability.

\section{Method}

As mentioned previously, pruning requires consideration of both feature information and gradient information. Inspired by the success of Seg-Grad-CAM and RWS in generating gradient activation maps within convolutional layers, we leverage Seg-Grad-CAM and RWS to identify channels in the original pre-trained convolutional network that provide strong support across all target classes, as shown in Fig. \ref{Fig2}. In the following sections, we provide a detailed description of our proposed Feature-Grad-Prune (\model) method.

\subsection{Channel Heatmap Generation}

The goal of computing the channel heatmap is to identify the significant channels that should be retained for preserving important channel information. Channels with high heatmap values indicate their importance to the task. \model utilizes the channels with high heatmap values to assess their relevance to a particular target class. Therefore, the first stage of \model is to compute the channel heatmaps for each convolutional layer.
                  
Given a trained Model, the method for calculating the channel heatmap for each convolutional layer is as follows:

\begin{equation}
    \centering	
     H^{d}_{j} = ReLU\left( \sum_{m} G^{m}_{j} \mathcal{F}^{m}_{j} \right), G^{m}_{j} = GAP\left( \frac{\partial \sum p^{d}_{uv}}{\partial \mathcal{F}^{m}_{j}} \right)
    \label{equ1}
\end{equation}

\begin{equation}
    \centering	
     M = H^{d}_{j} \quad (if\; p^{d} = True),
    \label{equ2}
\end{equation}

\noindent where, \(H^{d}_{j} \) represents the heatmap for class \( d \) in the \( j \)-th layer (where \( j = 1, 2, ..., l-1, l \)). \( \mathcal{F}^{m}_{j} \) is the feature map obtained from the \( m \)-th convolutional kernel in the \( j \)-th layer, and \( \text{GAP} \) stands for global average pooling. The term \( p^{d}_{uv} \) refers to the output related to class \( d \), while \( p^{d} \) is the logit for the class \( d \). The coefficient \( G^{m}_{d} \) indicates the contribution of the \( m \)-th convolution kernel, and the ReLU function is applied to preserve only positive activations. $M$ is the channel heatmap required by the \model.

\subsection{Channels Ranking and Selection}

In the second stage, for a single-object segmentation task, $M$ can be used to compute the next stage. However, for multi-object segmentation, the performance of different targets across different channels may vary. \model aims to retain channels that provide strong support for all targets during the pruning process. Specifically, the candidate pruning channel indices are obtained as follows.

\noindent \textbf{Heatmap Representation.} Let \( M_{c,d} \) in Eq. \ref{equ2} represent the heatmap of the \( c \)-th channel for the \( d \)-th class in a convolutional layer, where \( c \in \{1, 2, \dots, C\} \) is the index of the channel, and \( d \in \{1, 2, \dots, D\} \) is the index of the class, with \( C \) being the total number of channels and \( D \) the total number of classes. $M_{c,d} \in \mathbb{R}^{H \times W}$, where \( H \times W \) represents the spatial dimensions of the heatmap.

\noindent \textbf{Importance for Class $d$.} For each class \( d \), the importance of class \( d \) is measured by summing the heatmaps of all channels \( c \). Specifically, the importance of channel \( c \) for class \( d \) is defined as:

\begin{equation}
    \centering	
    I_{c,d} = \sum_{i=1}^{H} \sum_{j=1}^{W} M_{c,d}(i,j),
    \label{equ3}
\end{equation}

\noindent here, \( I_{c,d} \) represents the importance of channel \( c \) for class \( d \), which is the sum of all pixel values in the heatmap \( M_{c,d} \).

\noindent \textbf{Multi-Class Support.} To select channels that have high importance for multiple classes \( d \), the total importance of each channel \( c \) across all classes \( d \) is computed. Specifically, the total importance of channel \( c \) is defined as follows, where \( I_{c,d} \) shows the importance of channel \( c \) for class \( d \),

\begin{equation}
    \centering	
    I_c = \sum_{d} I_{c,d}.
    \label{equ4}
\end{equation}

\noindent \textbf{Retaining Important Channels.} \model aims to retain channels that significantly contribute to multiple classes. Therefore, the channels are sorted based on their total importance $I_c$, and the Top $k$ of channels are retained. First, all channels are sorted by their total importance $I_{c}$:

\begin{equation}
    \centering	
    I_{c(1)} \geq I_{c(2)} \geq \dots \geq I_{c(C)},
    \label{equ5}
\end{equation}

\noindent where \( c(1), c(2), \dots, c(C) \) are the channels sorted in descending order of \( I_c \). \model retain the Top $k$ channels.

\begin{equation}
    \centering	
    \mathcal{C}_{\text{retain}} = \{ c(1), c(2), \dots, c(\lceil k \cdot C \rceil) \},
    \label{equ6}
\end{equation}

\noindent here, \( k \) represents the percentage of channels to retain, and \( \lceil k \cdot C \rceil \) is the number of channels to be preserved based on this percentage. These channels have the highest total importance across all classes and are kept for further use. The remaining channels in the convolutional layers will be removed from the model. At this point, \model has identified the channels to be pruned.

\subsection{Model Pruning and Fine-tuning}
In the third stage, \model constructs the pruned model \( \text{prune\_model} \) based on the pruning information obtained in the second stage and loads the pruned weights. First, the model structure \( \text{New\_model} \) is rebuilt to accommodate the pruned channels. Then, the weights from the original model \( \text{model} \) are copied to the corresponding convolutional layers and channels in \( \text{New\_model} \). For each convolutional layer \( \text{Conv}_l \) and its retained channel set \( \mathcal{C}_{\text{retain}} \), the weights \( W_{l,i} \) and biases \( b_{l,i} \) from the corresponding channels in the original model are transferred to \( \text{New\_model} \). After several rounds of training and fine-tuning on \( \text{New\_model} \), \model obtains the final pruned model. It should be noted that \model currently performs pruning only on the convolutional layer channels, while the parameters of other layers, such as batch normalization and fully connected layers, are copied with the same \( \mathcal{C}_{\text{retain}} \) size. Once the pruned model is obtained, the pruned weights are loaded, and inference can be performed.

\section{Experiments}
\subsection{Experimental Setup}

\noindent \textbf{Datasets and Baselines.} To evaluate the effectiveness of the \model pruning algorithm, we designed a series of comparative experiments covering both image classification and segmentation tasks. These experiments aim to verify \model’s pruning efficacy and performance retention across various model architectures and datasets. For image classification, we used the CIFAR-10 (10 classes) and CIFAR-100 (100 classes) datasets, with VGGNet and ResNet-50 selected as baseline models to represent classic deep convolutional and residual network structures, effectively testing \model’s generalizability and adaptability. For segmentation tasks, we used the CamVid (32 classes) and Cityscapes (19 classes) datasets, employing SegNet and ResNet-50 as backbone networks respectively to demonstrate \model's effectiveness in complex feature learning scenarios.

\noindent \textbf{Evaluation Metrics.} To assess the size and computational demand of pruned models, commonly adopted evaluation metrics were used, including parameter count and FLOPs. For task-specific performance, classification tasks report the model's accuracy on CIFAR-10 and CIFAR-100, while for segmentation tasks, mIOU is used as the primary metric to evaluate the segmentation performance of pruned models.

\noindent \textbf{Training Configuration:} For classification tasks, networks were trained for 200 epochs using the SGD optimizer with an initial learning rate of 0.05, which decayed by a factor of 10 at epochs 100 and 150. The momentum was set to 0.9, weight decay to \(5 \times 10^{-4}\), and batch size to 128. For segmentation tasks, the DeepLabV3 model was adopted with both ResNet-50 and SegNet as backbone networks. Input images were randomly cropped to 512×512 pixels. Training also used the SGD optimizer, with an initial learning rate of 0.01, momentum of 0.9, and weight decay of \(10^{-4}\). The batch size was set to 16, and all experiments were conducted on a single NVIDIA Tesla 4090 GPU.

\noindent \textbf{Pruning Configuration:} During pruning, \model scores the importance of each channel and performs channel selection based on the optimal Top $k$ value (0.35 or 0.4) determined from ablation studies, retaining channels that are important for all classes. After pruning, the model undergoes fine-tuning to recover any potential performance loss. The fine-tuning process follows the original training configuration, maintaining the same learning rate schedule, optimizer (SGD) settings, and data augmentation strategies, with fine-tuning conducted for 10 epochs.

\subsection{Performance}

\noindent \textbf{Image Classification Task.} To comprehensively evaluate the proposed \model pruning method, we conducted extensive experiments on the CIFAR-10 and CIFAR-100 datasets using the VGG-16 and ResNet-50 models, respectively. The results, as shown in Table \ref{table1}, compare the FGP method with other mainstream pruning methods in terms of Top-$k=0.35/k=0.4$ accuracy, FLOPs reduction (PR), and parameter reduction (PR).

For the VGG-16 model on CIFAR-10, Table \ref{table1} presents the performance of various pruning methods, including the unpruned baseline, traditional methods such as L1, HRank, and FPGM, as well as our proposed FGP method. The results show that FGP exhibits significant advantages in Top-1 accuracy, FLOPs, and parameter reduction, effectively reducing model complexity while maintaining high accuracy. Specifically, when the pruning rate is 23.62\% ($k=0.35$, denoted as FGP$^\ast$), FGP$^\ast$ reduces FLOPs to 74.16M while maintaining an accuracy of 93.92\%, comparable to the unpruned model and superior to methods like HRank and GT. When the pruning rate increases to 26.48\% ($k=0.4$, denoted as FGP$^\dagger$), FGP$^\dagger$ further reduces parameters to 1.73M with an accuracy of 93.49\%, demonstrating the method's ability to reduce model complexity while maintaining high accuracy. This demonstrates the consistency of the FGP method across different datasets, efficiently preserving model performance while reducing computational resource requirements.

Similarly, we conducted pruning experiments on the ResNet-50 model on both CIFAR-10 and CIFAR-100 datasets. The results in Table 1 indicate that FGP also performs exceptionally well on the ResNet-50 architecture. FGP$^\ast$ achieves a pruning rate of 53.96\% on CIFAR-10 with an accuracy of 92.87\%, and maintains a high accuracy of 73.16\% on CIFAR-100. Compared with pruning methods such as ThinNet, L1, HRank, FPGM, and DCP-Adapt, FGP shows clear advantages in Top-1 accuracy and model complexity, significantly reducing FLOPs and parameter count while maintaining high accuracy. This demonstrates that the FGP method has broad applicability, adapting well to different network structures and task requirements, showcasing outstanding performance and efficiency across various tasks.

\begin{table*}[t]
\centering
\caption{Comparison of Pruning Methods on CIFAR-10 and CIFAR-100 with VGG-16 and ResNet-50 Backbone}
\normalsize
\renewcommand\arraystretch{1.09}

\begin{tabular}{c|ccccc}
\hline
\multirow{2}{*}{Backbone} & \multirow{2}{*}{Method} & \multirow{2}{*}{FLOPs (\%)} & \multirow{2}{*}{Parameters (\%)} & \multicolumn{2}{c}{Accuracy (\%)} \\ \cline{5-6} 
                              &                         &                             &                                  & CIFAR-10                               & CIFAR-100                               \\ \hline
\multirow{5}{*}{VGG-16}       & \textit{Unpruned}                &\textit{314.03M (0\%)}                             &\textit{17.7M (0\%)}                             &\textit{93.92}                                        &\textit{74.10}                                         \\
                              & Slimming \cite{liu2017learning}                     &162.90M (51.88)                    &5.4M (12.03)                         &92.91                                      &73.36                                     \\
                              & GT \cite{yu2021gate}                 &99.97M(31.82)                            &1.99M (13.52)                                  &93.27                                       &70.91                                        \\
                              & HRank \cite{lin2020hrank}                   &73.62M (23.50)                 &1.23M (8.00)                        &91.23                                      & 72.31                                     \\ \cline{2-6}
                              &\model$^{*}$             &74.16M (23.62)                            &1.47M (8.40)                                  &92.68                                        & 72.52                                        \\

                                & \model$^{\dagger}$            &83.15M (26.48)                            &1.73M (9.78)                                  &93.49                                        & 72.82                                        \\
                            
                             \hline
\multirow{5}{*}{ResNet-50}    & \textit{Unpruned}              &\textit{1.30G (0\%)}                            &\textit{23.52M (0\%)}                                  &\textit{95.09}                                        &\textit{78.60}                                         \\
                              & FPEI \cite{wang2021filter}               &646.94M(49.84)                             &13.64M (57.53)                                  &91.85                                        &69.58                                        \\
                              & SlimConv \cite{qiu2021slimconv}               &853.50M (64.12)               &14.83M (63.05)                     &93.12                                      &73.85                                      \\ \cline{2-6}   
                              & \model$^{*}$               & 718.26M (53.96)                            &13.76M (58.51)                                 &92.87                                        & 71.54 \\             
                              & \model$^{\dagger}$              & 751.84M (56.48)                            &14.03M (59.65)                                 &93.46                                        & 74.70 
                              \\ \hline
\end{tabular}
\label{table1}
\end{table*}

\noindent \textbf{Image Segmentation Task.} To further validate the effectiveness of the proposed FGP pruning method, we conducted image segmentation experiments on the CamVid and Cityscapes datasets. The experimental setup includes comparisons with general pruning algorithms (such as L1 norm pruning, ThiNet, and FPGM) and specialized segmentation pruning algorithms (such as CAP, MTP, and PDC). To ensure fairness, all \model's experiments are based on the Deeplabv3 model. The unpruned baseline models were directly trained under the same settings.

The experimental results in Table \ref{table2} show that, in terms of mIoU, FLOPs reduction, and parameter reduction in the segmentation tasks, the FGP method significantly reduces computational cost and parameter count while maintaining high segmentation accuracy. On the CamVid dataset using SegNet as the backbone, FGP$^\ast$ achieves a pruning rate of 30.58\%, reducing FLOPs to 32.64G and parameters to 8.71M, while achieving an mIoU of 53.61\%, outperforming methods like CAP and Slimming. When the pruning rate is further increased to 36.06\% (FGP$^\dagger$), the mIoU remains high at 54.94\%, demonstrating the method's robustness in maintaining performance even at higher pruning rates.

In the more challenging Cityscapes dataset with ResNet-50 as the backbone, the advantages of FGP are even more pronounced. FGP$^\ast$ achieves a pruning rate of 38.58\%, with an mIoU of 77.1\%, closely matching the unpruned model's 79.3\%. FGP$^\dagger$, with a pruning rate of 39.61\%, maintains an mIoU of 79.0\%, outperforming all other methods. These results indicate that the FGP method can not only adapt to different backbone structures (such as ResNet-50 and VGG-16) but also demonstrates superior performance and adaptability across various tasks and datasets.

\begin{table*}[t]
\centering
\caption{Comparison of Pruning Methods on CamVid and Cityscapes Datasets with SegNet and ResNet-50 Backbones}
\normalsize
\renewcommand\arraystretch{1.09}
\begin{tabular}{c|c|c c c c}
\hline
Dataset                     & Backbone                   & Method    & FLOPs (\%) & Parameters (\%) & mIoU (\%) \\ \hline
\multirow{5}{*}{CamVid}     & \multirow{5}{*}{SegNet}    & \textit{Unpruned}  &\textit{106.73G (0\%)}            &\textit{29.45M (0\%)}                 &\textit{55.6}           \\
                            &                            & CAP \cite{he2021cap}       &53.14G (49.79)            &11.47M (38.95)               &54.12           \\
                            &                            & Slimming \cite{liu2017learning}               &47.81G (45.80)            & 12.25M (41.60)                &54.78           \\
                            &                            & FPGM \cite{he2019filter}        &32.95G (30.87)      &15.63M (53.08)                             &52.54           \\ \cline{3-6}

                            &                            & \model$^{*}$ &\textbf{32.64G (30.58)}          &\textbf{8.71M (29.58)}                 &53.61           \\ 
                                                        &                            & \model $^{\dagger}$  & 38.49G (36.06)         & 10.82M (36.74)              & \textbf{54.94}          \\ 
                            \hline
\multirow{8}{*}{Cityscapes} & \multirow{8}{*}{ResNet-50} & \textit{Unpruned}  &\textit{1418.29G (0\%)}            &\textit{46.01M (0\%)}                 &\textit{79.3}           \\
                            &                            & Taylor \cite{molchanov2019importance}    &564.88 (39.83)            &14.99 (32.59)                 &78.3           \\
                            &                            & DepGraph \cite{fang2023depgraph}     &561.92G (39.61)            &16.84M (36.61)                 &76.6           \\ 
                            &                            & Slimming \cite{liu2017learning}               &558.72G (39.39)            &15.57M (33.85)                 &77.6      \\
                            &                            & DCFP \cite{wang2023dcfp}    &555.68G (39.18)                 &14.85M (32.28)   &78.8         \\
                            &                            & FPGM \cite{he2019filter}        &553.84G (39.05)            & 16.49M (35.58)                &76.5           \\ \cline{3-6}
                            &                            & \model$^{*}$ &\textbf{547.19G (38.58)}           &\textbf{14.42M (31.35)}                 &77.1           \\ 
                                                        &                            & \model $^{\dagger}$  & 561.72G (39.61)          & 16.27M (35.37)                & \textbf{79.0}          \\ 
                            \hline
\end{tabular}
\label{table2}
\end{table*}

\subsection{Parameter Analysis}

\noindent \textbf{Study on Top $k$.} In this experiment, we employed \model and conducted an ablation study on the CIFAR-10 dataset using the VGGNet-16 architecture. The main objective of the experiment was to explore the impact of different Top $k$ values on model performance in order to determine the optimal number of channels to retain, thereby optimizing both accuracy and inference speed while compressing the model. 

\begin{figure}[h]
    \centering
    \begin{subfigure}[t]{\columnwidth}
        \centering
        \includegraphics[width=\linewidth]{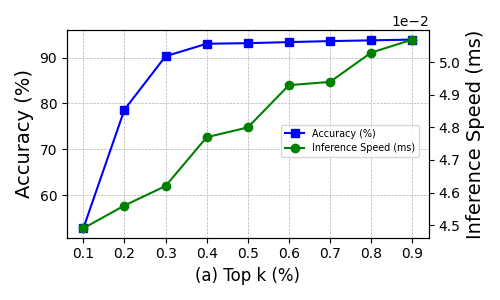}
        \label{fig:pruning_rate_accuracy}
    \end{subfigure}
    
    \vspace{-2em} 

    \begin{subfigure}[t]{\columnwidth}
        \centering
        \raisebox{2em}{ 
            \hspace{-0.08\columnwidth} 
            \includegraphics[width=0.9\linewidth]{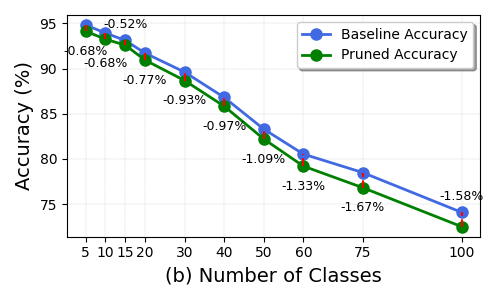} 
        }
        \label{fig:classes_accuracy}
    \end{subfigure}
    \vspace{-4em} 
    \caption{Parametric experiments with Top $k$ and classes}
    \label{Fig3}
\end{figure}

As shown in Fig. \ref{Fig3} (a), we tested Top $k$ values ranging from 10\% to 90\%, comparing changes in model accuracy and inference speed under different retained channel quantities.The experimental results indicate that as the Top $k$ value increases, i.e., as the number of retained channels increases, the model accuracy gradually improves within a certain range, and the inference speed also improves. However, when the Top $k$ value exceeds a certain threshold, the improvement in accuracy begins to plateau, while the inference speed starts to decline. This is because, near this threshold, the vast majority of important channels have already been retained, and further increasing the Top $k$ value only introduces channels with relatively low contribution to model performance, thus having a limited impact on accuracy improvement. At the same time, as the number of retained channels increases, the computational complexity also increases, resulting in a decrease in inference speed.

Notably, when the Top $k$ value is between 0.35 and 0.4, the model achieves an optimal balance between accuracy and inference speed. Increasing the Top $k$ value beyond this range may slightly improve accuracy but negatively impacts inference speed. Therefore, selecting a Top $k$ value between 0.35 and 0.4 can significantly enhance inference speed while maintaining model accuracy, achieving an optimal balance between model compression and performance optimization.In other experiments within this paper, the Top $k$ value is also set to either 0.35 or 0.4 to maintain this optimal balance.


\noindent \textbf{Study on number of Classes.} Since the \model method retains only channels with high support values for all classes during pruning, it is necessary to conduct an ablation study on the number of classes to evaluate the effectiveness of this method under varying class counts. This study helps us understand the applicability of \model in both few-class and many-class tasks, especially in terms of its impact on model performance while retaining high-support channels. The experiment is based on the CIFAR-100 dataset, using the VGGNet-16 model, and was conducted by controlling the number of classes in the dataset (set to 5, 10, 15, 20, 30, 50, 75, and 100 classes), with the Top $k$ value fixed at 0.4.

As shown in Fig. \ref{Fig3} (b), it can be observed that as the number of classes increases, the accuracy gap between the pruned model and the baseline model gradually widens. For tasks with a small number of classes (such as 5, 10, and 15 classes), the accuracy gap between the pruned model and the baseline model remains relatively small, at 0.68\%, 0.68\%, and 0.52\%, respectively. This indicates that in tasks with fewer classes, the \model method can effectively retain channels with high support values for these classes, thereby achieving model compression and inference speedup while maintaining a high accuracy close to the baseline.

However, as the number of classes increases to 100, the accuracy gap between the pruned model and the baseline model increases to 1.58\%. This trend reflects the characteristic of the \model method—since it retains only channels with high support values for all classes, when the number of classes increases, the retained channels cannot fully capture the complex feature representations required for a large number of classes, resulting in a more significant drop in accuracy.

\subsection{Ablation Study}

\noindent \textbf{Different Selection Criteria.} In this ablation study, we used the VGG-16 model on the CIFAR-10 dataset to compare the effectiveness of three different pruning selection criteria: feature-based pruning, gradient-based pruning, and our proposed \model method, which integrates both feature and gradient information. Each pruning method was applied with the same pruning rate, and the results are shown in Table \ref{table3}. 
The feature-based pruning method (AOFP) relies primarily on feature information, reducing the model’s FLOPs from 314.03M to 97.52M, with a significant decrease in accuracy to 93.30\%. While feature-based pruning effectively reduces computational cost, it lead to a notable accuracy drop for that important feature information is lost. Meanwhile, the gradient-based pruning method (GFBS), which focuses on gradient information, achieves a 70.00\% reduction in FLOPs, reducing FLOPs to 94.21M. However, this method neglects feature information, resulting in accuracy drop to 92.79\%, indicating that relying solely on gradient information may sacrifice model generalization.

Our proposed \model method integrates both feature and gradient information to comprehensively evaluate channel importance, effectively identifying channels crucial to model performance. As a result, \model achieves a reduction in FLOPs to 83.15M (a 73.52\% reduction) while maintaining a high accuracy of 93.49\%. Compared to methods that rely solely on feature or gradient information, \model achieves a more desirable balance between model compression and accuracy retention, demonstrating its superiority as an efficient pruning strategy.

\begin{table}[h]
    \centering
    \caption{Comparison of different selection criteria on VGG-16 for CIFAR-10: Feature-Based, Gradient-Based, and FGP Methods }
    \resizebox{\columnwidth}{!}{
        \begin{tabular}{c c c} 
        \toprule
        Pruning Methods & FLOPs ↓ (\%) & Acc (\%) \\
        \midrule
        \textit{Unpruned} & \textit{314.03M} & \textit{93.92} \\
        AOFP (Feature-based) \cite{ding2019approximated}  & 97.52M (70.81) & 93.08 \\
        GFBS (Gradient-based) \cite{liu2023generalized} & 94.21M (70.00) & 92.79 \\ 
        \model$^{\dagger}$ &83.15M(73.52) &93.49 \\
        \bottomrule
        \end{tabular}
    }
\label{table3}
\end{table}

\noindent \textbf{Channel Retention Configurations.} In this ablation study, we conducted pruning experiments on VGG-16 and ResNet-50 using the CIFAR-10 dataset to explore the advantages of retaining the Top k channels based on channel contribution, rather than using a fixed number of channels for pruning. In the \model method, we refer to the proportion of channel contributions when retaining channels, selecting channels whose cumulative contributions reach a predefined threshold $k$. This dynamic determination of retained channels avoids the limitations associated with fixed-number pruning.

In the experiment, we compared various pruning configurations, including Peak k, Peak 30/50 Rand, Bot 30/50 Rand, and Rand 30\%. Specifically, Peak k retains the top-ranked channels, while Peak 30/50 Rand and Bot 30/50 Rand denote randomly selecting 30\% of channels from the top 50\% and bottom 50\% channels, respectively. Rand 30\% randomly selects 30\% of the channels for retention. Through these comparative experiments, we evaluated the impact of different pruning strategies on model performance.

\begin{figure}[h]
    \centering
    \includegraphics[width=\columnwidth]{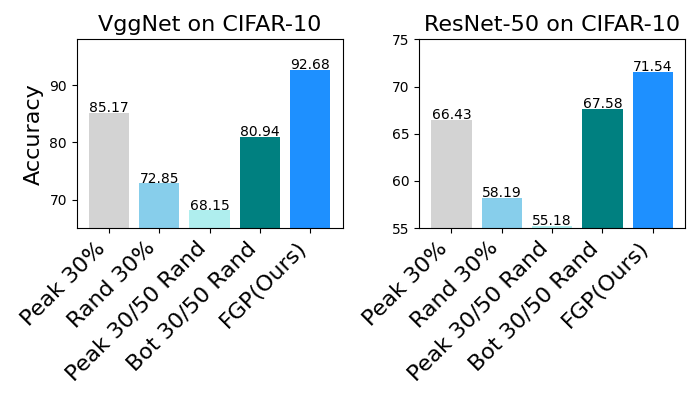}
    \caption{Accuracy for different channel retention configurations} 
    \label{Fig4}
    \vspace{-1em} 
\end{figure}

As shown in Fig. \ref{Fig4}, the experimental results indicate that the \model method outperforms other pruning strategies on both VGG-16 and ResNet-50, achieving accuracies of 92.60\% and 71.54\%, respectively. In contrast, pruning strategies that retain a fixed number or randomly selected channels (e.g., Peak 30\% and Rand 30\%) yielded lower accuracy, with VGGNet achieving only 85.17\% and ResNet-50 reaching 66.43\%. These results validate the effectiveness of our proposed channel importance ranking method and demonstrate that the Top k strategy, which considers the proportion of contributions during channel retention, can more effectively identify critical channels, balancing model compression and performance retention.

\section{Conclusion}

In this study, we propose the \model method to enhance lightweight model performance through pruning based on feature and gradient information from convolutional layers. \model outperforms existing methods across multiple datasets, ensuring that the pruned model retains high performance by selecting the most impactful channels. Future work should explore the effects of non-convolutional layer structures on pruning and assess \model's adaptability across different tasks, such as classification, detection, and segmentation. Overall, \model effectively prunes convolutional models while maintaining performance.

{
    \small
    \bibliographystyle{ieeenat_fullname}
    \bibliography{main}
}

\end{document}